\pdfoutput=1

\documentclass[11pt]{article}

\usepackage[final]{acl}

\usepackage{times}
\usepackage{latexsym}
\usepackage{graphics}
\usepackage{booktabs}
\usepackage{tabularx}
\usepackage{makecell}
\usepackage{multirow}
\usepackage{float}
\usepackage{algorithm}
\usepackage{algpseudocode}
\usepackage{amsmath}

\algrenewcommand\algorithmiccomment[1]{\textcolor{blue}{// \textit{#1}}}
\newenvironment{algorithmfont}{\fontsize{9.5pt}{12pt}\selectfont}{\par}

\usepackage{subcaption}
\usepackage[T1]{fontenc}

\usepackage[utf8]{inputenc}

\usepackage{microtype}
\usepackage{inconsolata}

\usepackage{graphicx}
\usepackage{amssymb}
\usepackage{pifont}
\newcommand{\cmark}{\ding{51}}%
\newcommand{\xmark}{\ding{55}}%
\usepackage{todonotes}[colorinlistoftodos,prependcaption,textsize=tiny]

\usepackage{amsmath}
\usepackage{enumitem}
\title{S2D: Sorted Speculative Decoding For More Efficient Deployment of Nested Large Language Models}

\author{
Parsa Kavehzadeh$^2$ \And  Mohammadreza Pourreza$^2$ \And   Mojtaba Valipour$^{1}$ \\ \AND
Tianshu Zhu$^2$ \And Haoli Bai$^2$ \And Ali Ghodsi$^1$ \\ \AND Boxing Chen$^2$ \And  Mehdi Rezagholizadeh$^2$ \\
        \AND 
        \small{\textmd{\{mojtaba.valipour, ali.ghodsi\}@uwaterloo.ca, \{mehdi.rezagholizadeh, parsa.kavehzadeh\}@huawei.com}}
        \AND 
        \textmd{1: University of Waterloo, 2: Huawei Noah’s Ark Lab}
}

\begin{document}
\maketitle
\begin{abstract}


Deployment of autoregressive large language models (LLMs) is costly, and as these models increase in size, the associated costs will become even more considerable. Consequently, different methods have been proposed to accelerate the token generation process and reduce costs. Speculative decoding (SD) is among the most promising approaches to speed up the LLM decoding process by verifying multiple tokens in parallel and using an auxiliary smaller draft model to generate the possible tokens. In SD, usually one draft model is used to serve a specific target model; however, in practice, LLMs are diverse, and we might need to deal with many target models or more than one target model simultaneously. In this scenario, it is not clear which draft model should be used for which target model, and searching among different draft models, or training customized draft models, can further increase deployment costs. In this paper, we first introduce a novel multi-target scenario for deployment of draft models for faster inference. Then, we present a novel more efficient sorted speculative decoding mechanism that outperforms regular baselines in multi-target setting. We evaluated our method on Spec-Bench in different settings including base models such as Vicuna 7B, 13B, and LLama Chat 70B. Our results suggest that our draft models perform better than baselines for multiple target models at the same time. 




\end{abstract}

\section{Introduction}
Large language models (LLMs) have advanced very quickly and become popular in different academic and industrial domains \cite{brown2020language}. As the size of these models increases \cite{narayanan2021efficient}, accelerated inference is becoming more popular to reduce the overhead costs of their deployment. There are an increasing number of publications in the literature trying to achieve faster inference~\cite{stern2018blockwise, chen2023accelerating, leviathan2023fast, chen2023cascade}. These different approaches include, but are not limited to, reducing redundant layers~\cite{men2024shortgpt}, quantization~, early exiting~\cite{varshney2023accelerating}, optimizing the KV-caching of transformers~\cite{zhang2023h2o}, and speculative decoding~\cite{leviathan_fast_2023,chen2023accelerating}. In this paper, we focus on speculative decoding (SD) as one of the most prominent solutions (due to its simplicity and widespread usage) for improving the decoding speed of LLMs.

SD is a technique based on drafting and verification. Since autoregressive generation of LLMs is a sequential process, SD tries to use a faster proxy model to generate a candidate draft (with a fixed pre-defined length). Then, the given generated tokens by the draft model will be sent to the target LLM in one forward pass to be verified. In regular SD, the draft model is a separate smaller language model; however, we have \textit{self-speculative} solutions~\cite{zhang2023draft, elhoushi2024layer, zhong2024s3d} in which the draft model is a part of the target model. 

While SD is pretty popular in the literature and we have many of its variants available, there are a few bottlenecks in SD, which we will focus on in our paper: \textbf{1-``search problem''} we can have target models with different sizes and it is not clear how to obtain the proper draft model for each target model.
Moreover, target models can be trained on different downstream tasks, and using a single draft model to serve all the tasks might not yield the best results. This might lead to a distribution mismatch between the target and the draft model unless both the target and the draft model are updated. 
\textbf{2- ``minimal training''} We prefer not to train or modify the target model received from the users. This means that most of the solutions in the category of self-speculative solutions are not within our scope.  

To address the mentioned problems, we propose our solution called \textit{sorted speculative decoding} (S2D). Sorted refers to the \textit{sorted-training}~\cite{valipour2023sortednet} approach in which a model and its selected sub-models can be trained on single or multiple tasks at the same time. Inspired by sorted training, our S2D trains multiple draft models in one model to be able to serve more than just one target model at a time (without needing to maintain multiple draft models) to take care of the \textit{search problem}. In this regard, the initial draft model is extracted from the target model and after designing the sub-models, they are trained together. Moreover, in contrast to the self-speculative solutions, our approach is just applied to the draft side and we do not need to train the target model. Finally, to make an efficient use of the trained sorted draft models, we use an adaptive draft selection mechanism. 

The contributions of this paper are listed as follows: 1-  \textbf{Introduction of Multi-Target Draft Models}: We pioneer the concept of employing a single draft model that can simultaneously accommodate multiple target models, reducing deployment complexity and costs. 2- \textbf{Development of a Sorted Speculative Decoding Mechanism}: Our S2D mechanism leverages sorted fine-tuning, enabling the creation of sub-models within a draft model, without the necessity of maintaining separate draft models for each target LLM. 
3- \textbf{Adaptive Draft Selection Strategy}: We introduce an adaptive draft selection mechanism that optimally chooses sub-models based on confidence thresholds. 4- \textbf{Comprehensive Evaluation on Spec-Bench}: We rigorously evaluated our S2D method on the Spec-Bench.



\section{Related Work}

For LLMs to perform more efficiently, efficient decoding/sampling methods \cite{leviathan2023fast, li2024eagle} are essential. As LLMs grow in complexity and size, the need for innovative techniques to enhance their speed and accuracy becomes even more pressing. Various methodologies are discussed in this literature review, including parallel sampling, speculative decoding, and early exit strategies, alongside their contributions and advancements.

\paragraph{Parallel Decoding}


The first mechanism aiming to accelerate the inference process of large language models was presented in \cite{stern2018blockwise}. This paper introduced a blockwise parallel decoding strategy, aiming to generate the next k tokens simultaneously in a single forward pass using a set of auxiliary models. Then they proposed to use the same language model to verify the generated tokens in parallel. This simple draft-then-verify mechanism, as discussed in the paper, can potentially reduce the number of forward passes from m to m/k+1 \cite{stern2018blockwise}. 

\paragraph{Speculative Decoding}

Inspired by \cite{sun2021instantaneous}, \cite{xia2022lossless} proposed a draft-then-verify mechanism to aggressively generate a fixed number of tokens in parallel without the new tokens depending on the previous ones, and then verify the generated tokens in one forward pass. 

Later, they \cite{xia2023speculative} proposed a more advanced attention mechanism to generate independent tokens in parallel by using distinct attention queries instead of using a shared attention query, or simply adding more language model heads as done in the past. In addition, for the verification process, they also relaxed the top-1 greedy decoding. Instead, they proposed to accept any token from the top-$\beta$ candidates as long as their score gap is not far from the most likely token \cite{xia2023speculative}.


\paragraph{Speculative Sampling}

Other methods, such as \cite{chen2023accelerating}, \cite{leviathan2023fast}, generalized speculative decoding to the stochastic non-greedy setting. As these methods are just a variant of the draft-then-verify mechanism with a modified rejection sampling algorithm for ensuring sampling quality, we will leave the integration of these methods with our proposed method as future work. 

\paragraph{Self-Speculative Decoding}
Other methods, such as \cite{zhang2023draft}, introduced self-speculating, which tries to get rid of the auxiliary models by selectively skipping certain intermediate layers during the drafting phase. As we can use the full LLM to validate the generated tokens, without any additional model we can enjoy accelerated inference. This is also aligned with approaches like \cite{elhoushi2024layer}, \cite{chataoui2023jointly}, \cite{kavehzadeh2024sorted}, and \cite{valipour2023sortednet}.

\paragraph{Other Methods}

More recently, new techniques \cite{liu2023online, chen2023cascade, li2024eagle, cai2024medusa, fu2024break, NEURIPS2023_6034a661, miao2023specinfer, varshney2023accelerating, ankner2024hydra, he2023rest, yi2024generation} have emerged that trying to incorporate sophisticated mechanisms to further improve the speculative sampling speedup gain. 


For simplicity, this paper will focus on the Self-Speculative Decoding setting, but our method is also applicable to other speculative sampling methods with minor adjustments, without loss of generality.

\paragraph{Benchmarks}

In addition, to evaluate these different algorithms, several benchmarks \cite{zheng2024judging, taori2023alpaca, chen2021evaluating} can be used to measure performance and speed up gains. One of the most comprehensive benchmarks, however, is Spec-Bench, specifically designed to evaluate speculative decoding methods \cite{xia2024unlocking}. Spec-Bench comprises 6 sub-tasks: translation, multi-turn conversation (MT-Bench), retrieval-augmented generation, mathematical reasoning, question answering, and summarization, each with 80 instances. In this paper, we will focus mainly on Spec-Bench. 

\section{Methodology}

\subsection{Background}

\paragraph{Speculative Decoding:} Speculative decoding is a two-step process involving drafting and verification. At each decoding step, a draft model efficiently generates multiple potential future tokens, which are then verified in parallel by the target model at inference time. Specifically, during the drafting step, given an input sequence $\{x_1, \ldots, x_n\}$ and the target LLM $M_t$, a faster drafter model $M_d$ decodes the next $K$ drafted tokens as a speculation of the target LLM's output \cite{xia2024unlocking}:
\begin{equation}
\begin{aligned}
p_1, \ldots, p_K &= \text{DS}(x \leq k, M_d), \\
& \hat{x_i} \sim p_i, \quad i = 1, \ldots, K
\end{aligned}
\end{equation}

where DS(·) represents the drafting strategy, \( p \) is the conditional probability distribution calculated by \( M_d \), and \( \hat{x_i} \) is the token sampled from the draft model's probability distribution \( p_i \). The tokens generated by the draft model are then verified by the target LLM \( M_t \). Given the input sequence \( \{x_1, \ldots, x_n\} \) and the drafted tokens \( \{\hat{x_1}, \ldots, \hat{x_K}\} \), the \( M_t \) model is used to measure \( K + 1 \) probability distributions simultaneously as follows:
\begin{equation}
q_i = M_t(x \mid x \leq t, x_{<i}), \quad i = 1, \ldots, K + 1
\end{equation}

Each drafted token \( \hat{x_i} \) is then verified using a specific verification criterion using \( \hat{x_i} \), \( q_i \), and \( p_i \). Only the tokens that meet this criterion are selected as final outputs.

\paragraph{Sorted Fine-tuning:} Sorted Fine-tuning \cite{valipour2023sortednet, kavehzadeh2024sorted} is a recently proposed approach for training many-in-one models by forming sub-models from a larger model. In the case of LLMs, sub-models are the sub-layers of the existing LLM. Each sub-model's output is predicted using the shared output prediction head from the last layer (original LLM head). To train the network, we define the loss as the summation of the losses of all the sub-models:
\begin{equation}
L = \frac{\sum_{n \in B} L_n(x; \theta_n)}{|B|}
\end{equation}

where \( L_n(x; \theta_n) \) is the loss for the \( n \)-th sub-model for input batch \( x \) and B denotes the number of sub-models.

\begin{algorithm}[t]
\caption{Sorted Speculative Decoding}
\begin{algorithmfont}
\begin{algorithmic}[1]
\Require Sorted draft layers $L$, Target model $f(\theta_{N})$, Input context $C$, Draft thresholds $T$, Draft candidates verification function VerifyTokens
\Ensure Generated sequence $S$

\Function{GenerateCandidates}{$S$}
    \State $candidates \leftarrow [\ ]$
    \While {not end of draft generation}
        \State \Comment{Adaptively generate draft candidates}
        \For {$n, threshold$ in zip$(L, T)$}
            \State $p_S \leftarrow f(S;\theta_{n})$
            \State \Comment{Sample from draft sub-model distribution}
            \State $x, c \sim p_S$ 
            \If {$threshold \leq c$}
                \State \textbf{append} $(x, c)$ \textbf{to} $candidates$
                \State \textbf{break} \Comment{Exit from intermediate draft}
            \EndIf
        \EndFor
    \EndWhile
    \State \Return $candidates$
\EndFunction

\State Initialize $S \leftarrow C$
\While {not end of sequence} \Comment{Initialize generation}
\State \Comment{Draft generation}
\State $Cands \leftarrow$ \Call{GenerateCandidates}{$S$}

\State \Comment{Verify draft tokens}
\State $Matches \leftarrow \text{VerifyTokens}(f(\theta_{N}), S, Cands)$
\State \textbf{append} $Matches$ \textbf{to} $S$
\EndWhile
\State \Return $S$

\end{algorithmic}
\end{algorithmfont}
\label{algorithm}
\end{algorithm}

\begin{figure*}[!t]
    \centering
    \scalebox{0.77}{\includegraphics{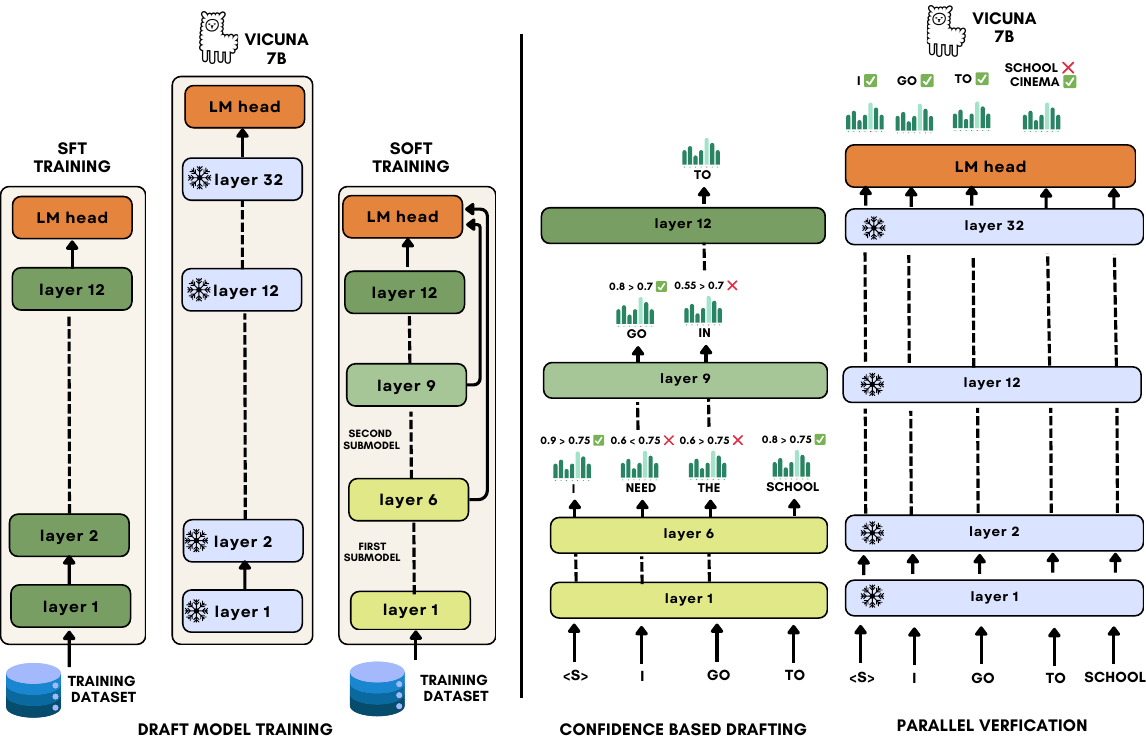}}
    \caption{\small{The figure on the left illustrates the draft model training process, comparing supervised fine-tuning (SFT) with Sorted fine-tuning (SoFT) using two sub-models with 6 and 9 layers. The figure on the right demonstrates the confidence-based drafting process, where the SoFT draft model is utilized to generate candidate tokens. The confidence thresholds for the two sub-models are set at 0.75 and 0.7, respectively.}
    }
    \label{fig:methodology}
\end{figure*}

\subsection{Why sorted draft instead of sorted target?}
\label{section:sorted_draft_vs_sorted_target}

In this paper, we introduce a method that involves Sorted fine-tuning (SoFT) of a draft model and using sub-models for sorted speculative decoding to increase the inference speed of multiple target models. An alternative method, is to use SoFT to train the target model, instead of the draft model, similar to the approach proposed in \cite{kavehzadeh2024sorted}. To evaluate these two methods, we fine-tuned the Llama2 13B \cite{touvron2023llama} on the GSM8K dataset using both standard supervised fine-tuning (SFT) and Sorted fine-tuning (SoFT) as described in \cite{kavehzadeh2024sorted}. The results are provided in Table \ref{tab:sorted_target_vs_sorted_draft}, where we compare our sorted speculative decoding with sorted draft model training with self sorted speculative decoding.

According to Table \ref{tab:sorted_target_vs_sorted_draft}, the sorted target model training method has three significant disadvantages. Firstly, it decreases accuracy by 16\% in final task performance and offers lower speed improvements because the sub-models used are larger than those in the sorted draft model training. Secondly, this method is not suitable for scenarios with multiple target models as it requires each target model to undergo SoFT training for self-speculative decoding to be applicable. Lastly, SoFT training of the target model incurs considerably higher costs compared to our method of SoFT training a smaller draft model.

\begin{table}
\setlength\extrarowheight{2pt}
\centering
\scalebox{0.57}{
\begin{tabular}{l|cc}
\toprule
& \multicolumn{2}{c}{\textbf{GSM8K}} \\
\midrule
\multirow{2}{*}{\textbf{Model}} & \multicolumn{2}{c}{\textbf{Auto-regressive Decoding}} \\ 
 & \textbf{Speedup} & \textbf{Accuracy} \\
\midrule
SFT (Llama2 13B) & 1$\times$ & 48.97 \\
\midrule
\multirow{2}{*}{\textbf{Model}} & \multicolumn{2}{c}{\textbf{Self Sorted speculative decoding (Sorted Target)}} \\
& \textbf{Speedup} & \textbf{Accuracy} \\
\midrule
Layers 12:40 (SoFT) & 1.21$\times$ & 33.51 \\
\midrule
\multirow{2}{*}{\textbf{Draft Model}} & \multicolumn{2}{c}{\textbf{Sorted Speculative Decoding (Sorted Draft)}} \\ 
 & \textbf{Speedup} & \textbf{Accuracy} \\
\midrule
Layer 6:12 (SoFT 6,9,12 13B) & \textbf{1.53$\times$} & 48.97 \\
\bottomrule
\end{tabular}
}
\caption{\small{Performance comparison between self sorted speculative decoding (sorted target) and adaptive speculative sampling (sorted draft) proposed in this paper on GSM8K dataset.}}
\label{tab:sorted_target_vs_sorted_draft}
\end{table}

\subsection{Sorted speculative decoding}
In this section, we introduce our approach which utilizes multiple draft models in the same architecture in an adaptive way to address multi-target inference acceleration problem. To reach this goal, we first introduce a new sorted draft architecture that can incorporate multiple draft sub-models in the same architecture. Then we explain the adaptive draft generation algorithm that we devise in order to use the draft sub-models efficiently in speculative decoding paradigm.

\paragraph{Training SoFT Draft}     
Supposed we have a pre-trained large language model $f(x;\theta_N)$ with the parameters $\theta$, input $x$ and $N$ number of layers. Also consider $f(\theta_n)$ as the sub-model with the parameters of first $n$ layers of the LLM ($n\leq{N}$). To reach our draft architecture, we first extract a sub-model with $f(\theta_{N_d})$, where $N_d<N$. Then we also determine three different sub-models in the extracted draft architecture as $f(x;\theta_{N_{ds}})$, $f(x;\theta_{N_{dm}})$ and $f(x;\theta_{N_{d}})$, where $N_{ds}<N_{dm}<N_d$. We utilize the sorted fine-tuning approach \cite{kavehzadeh2024sorted} to fine-tune the whole draft on the downstream dataset to reach three draft models with different sizes in the same architecture. In this paper, we use Vicuna 7B as the pre-trained language model with 32 layers. To define our draft sub-models, we set $N_{ds}$ to 6, $N_{dm}$ to 9, and $N_d$ to 12 in our experiments. Figure \ref{fig:methodology} (Left) shows the two SFT and SoFT methods to train an extracted draft model from the target Vicuna 7B.

\paragraph{Draft Generation} In order to make most out of the speculative decoding algorithm, we need to have both low latency draft models and high acceptance ratio compared to target model. To generate each token, we employ a confidence-based early-exiting approach in $f(x;\theta_{N_d})$ architecture. Supposed 
the draft sub-model layers $L_D = \{N_{ds}, N_{dm}, N_d\}$, we have the set of confidence thresholds $T_D = \{\tau_{ds}, \tau_{dm}, \tau_d\}$. To generate one draft token given input sequence $S$, we start iterating over draft sub-models, starting from $N_{ds}$. For each sub-model $N_i \in L_D$, we have:
\begin{equation}
    t, c \sim f(S;\theta_{N_i})
\end{equation}
Where $t$ and $c$ are the draft token and its confidence sampled from draft sub-model $N_i$. We accept the token $t$ as the final draft token if $c \geq \tau_{i}$. 
Algorithm \ref{algorithm} explains the draft generation mechanism of S2D algorithm in more details. Figure \ref{fig:methodology} (Right) also shows how draft generation works in S2D algorithm. 
 

\section{Experiments}

\subsection{Experimental Setup}
We selected the first 12 layers of Vicuna 7b checkpoint to build the architecture of our draft model. Then, we trained the draft model in both SFT and SoFT paradigms on the ShareGPT \footnote{\url{https://huggingface.co/datasets/anon8231489123/ShareGPT_Vicuna_unfiltered}} dataset for 3 epochs. We used the Spec-bench \cite{xia2024unlocking}, which is a benchmark for speculative decoding-based methods, to evaluate our draft models and the S2D algorithm. More details about the experimental setup and hyperparameters can be found in the Appendix \ref{appx:hyperparameters}.

\subsection{Baselines}
We categorize the baselines based on the dependency of draft training procedure on target models:

\paragraph{Target-Dependent Baselines}

\begin{itemize}[leftmargin=*]
    \item \textbf{Eagle \cite{li2024eagle}}: An approach proposing a single layer draft model trained with two feature alignment and cross-entropy losses based on target output.
    \item \textbf{Medusa \cite{cai2024medusa}}: A method for generating multiple draft candidates for future tokens by training multiple language model heads for each future token position.
    \item \textbf{Hydra \cite{ankner2024hydra}} A draft model based on recurrent neural architectures on top of the target model, which generates multiple draft candidates.
\end{itemize}

\paragraph{Target-Independent Baselines}
We have different scenarios for our draft generation as baselines:
\begin{itemize}[leftmargin=*]
    \item \textbf{SFT Checkpoint + Speculative Decoding}: We use the SFT checkpoint of first 12 Vicuna 7b layers as draft model in the speculative sampling algorithm.
    \item \textbf{Small sub-model of SoFT Checkpoint (6 Layers) + Speculative Decoding}: We use the smallest sub-model (Layer 6) of the SoFT checkpoint of first 12 Vicuna 7b layers as draft model in the speculative decoding algorithm.
    \item \textbf{Medium sub-model of SoFT Checkpoint (9 Layers) + Speculative Decoding}: We use the medium sub-model (Layer 9) of the SoFT checkpoint of first 12 Vicuna 7b layers as draft model in the speculative decoding algorithm.
    \item \textbf{Full SoFT Checkpoint (12 Layers) + Speculative Decoding}: We use the full SoFT checkpoint of first 12 Vicuna 7b layers as draft model in the speculative decoding algorithm.
    \item \textbf{Full SoFT Checkpoint (12 Layers) + S2D}: We use the full SoFT checkpoint of first 12 Vicuna 7b layers as draft model in the S2D algorithm. We set the thresholds of the intermediate sub-models from the ablations in section \ref{sec:thresholds}.
\end{itemize}

\subsection{Results}
In this section, we will discuss the results of the experiments we conducted to evaluate our SoFT draft model and S2D approach compared to other baselines.

\paragraph{Multi-Target Draft:} Table \ref{tab:results} shows the speedup ratio (compared to regular autoregressive inference of the target model) and mean accepted tokens length of different baselines on MT-Bench dataset in Spec-Bench. Fine-tuning an extracted network from Vicuna 7b, can result in speed-up in multiple targets with various sizes without any need for pre-training. Our S2D method outperforms or preserves the performance of the normal speculative decoding almost in all the target sizes. In smaller targets like Vicuna 7b where the latency of the draft is mostly important, S2D can adjust the draft generation procedure to be less slow by choosing the intermediate layers more likely. While in larger targets like LLaMA Chat 70b, it is shown that the capacity (mean accepted tokens length) is more important than the draft latency since 12 layer drafts gains higher speedups compared to the layer 6 of SoFT draft checkpoint. Even in this scenario S2D can maintain the optimum performance by adjusting the exiting layers accordingly.

\begin{table*}[t]
 \setlength\extrarowheight{2pt}
 \centering
 \scalebox{0.64}{\begin{tabular}{l|cc|cc|cc|cc|cc|cc|c}
  
  \toprule
   \multirow{3}{*}{\textbf{Method}} & \multicolumn{6}{c}{\textbf{Greedy (T = 0)}} & \multicolumn{6}{|c|}{\textbf{Non-Greedy (T = 1)}} &  \multirow{3}{*}{\textbf{\makecell[c]{Avg\\Speedup}}}
   \\ \cmidrule{2-13}
    & \multicolumn{2}{c}{\textbf{Vicuna 7B}} & \multicolumn{2}{c}{\textbf{Vicuna 13B}} & \multicolumn{2}{c|}{\textbf{LLaMA Chat 70B}} & \multicolumn{2}{c}{\textbf{Vicuna 7B}} & \multicolumn{2}{c}{\textbf{Vicuna 13B}} & \multicolumn{2}{c|}{\textbf{LLaMA Chat 70B}} &
   \\ 
    & Speedup & MAT & Speedup & MAT & Speedup & MAT & Speedup & MAT & Speedup & MAT & Speedup & MAT &
   \\ \midrule
   Eagle \cite{li2024eagle} & 2.62$\times$ & 3.84 & \xmark & \xmark & \xmark & \xmark & 2.05$\times$ & 3.42 & \xmark & \xmark & \xmark & \xmark & N/A\\
   Eagle (No Attention Tree) & 2.04$\times$ & 3.11 & \xmark & \xmark & \xmark & \xmark & 1.72$\times$ & 2.73 & \xmark & \xmark & \xmark & \xmark & N/A \\
   \midrule
   Medusa \cite{cai2024medusa} & 1.74$\times$ & 2.51 & \xmark & \xmark & \xmark & \xmark & 1.93$\times$ & 2.80 & \xmark & \xmark & \xmark & \xmark & N/A \\
   Medusa (No Attention Tree) & 1.34$\times$ & 1.76 & \xmark & \xmark & \xmark & \xmark & 1.48$\times$ & 1.92 & \xmark & \xmark & \xmark & \xmark & N/A \\
   \midrule
   
   Hydra \cite{ankner2024hydra} & 2.14$\times$ & 2.70 & \xmark & \xmark & \xmark & \xmark & 2.36$\times$ & 4.01 & \xmark & \xmark & \xmark & \xmark & N/A \\
   Hydra (No Attention Tree) & 1.78$\times$ & 2.65 & \xmark & \xmark & \xmark & \xmark & 2.03$\times$ & 3.11 & \xmark & \xmark & \xmark & \xmark & N/A \\
   \midrule

   
   SFT + SD & 1.19$\times$ & 3.19 & 1.21$\times$ & 3.05 & 1.94$\times$ & 2.46 & 1.16$\times$ & 3.44 & 1.10$\times$ & 3.16 & 1.94$\times$ & 2.54 & 1.42
   \\
   SoFT L6 + SD & \textbf{1.38$\times$} & 2.43 & 1.38$\times$ & 2.40 & 1.83$\times$ & 2.05 & \textbf{1.30$\times$} & 2.53 & 1.35$\times$ & 2.87 & 1.87$\times$ & 2.14 & 1.51
   \\
   SoFT L9 + SD & 1.24$\times$ & 2.80 & 1.31$\times$ & 2.78 & 1.92$\times$ & 2.27 &  1.26$\times$ & 3.05 & 1.32$\times$ & 2.87 & 1.94$\times$ & 2.34 & 1.49
   \\ 
   SoFT L12 + SD & 1.17$\times$ & 3.11 & 1.23$\times$ & 3.01 & 1.91$\times$ & 2.39 & 1.07$\times$ & 3.22 & 1.20$\times$ & 3.17 & 1.96$\times$ & 2.53 & 1.42
   \\
   SoFT + S2D (ours) & 1.34$\times$ & 2.86 & \textbf{1.38$\times$} & 2.76 & \textbf{1.95$\times$} & 2.36 & 1.27$\times$ & 3.01 & \textbf{1.38$\times$} & 2.89 & \textbf{1.98$\times$} & 2.44 & \textbf{1.55}
   \\
   \bottomrule
 \end{tabular}}
 \caption{
  \small{Overall Speedup and Mean Accepted Tokens lenght (MAT) on MT-Bench dataset in Multi-target inference acceleration setup. The speedups reported for Eagle, Medusa and Hydra are based on the publicly available checkpoints of these methods for Vicuna 7b target model and since these checkpoints cannot be applied to other targets, we denoted that by \xmark\ in the table.}
 }
 \label{tab:results}
\end{table*}

\begin{figure*}[t] 
    \begin{subfigure}{0.32\textwidth}
        \includegraphics[width=\linewidth]{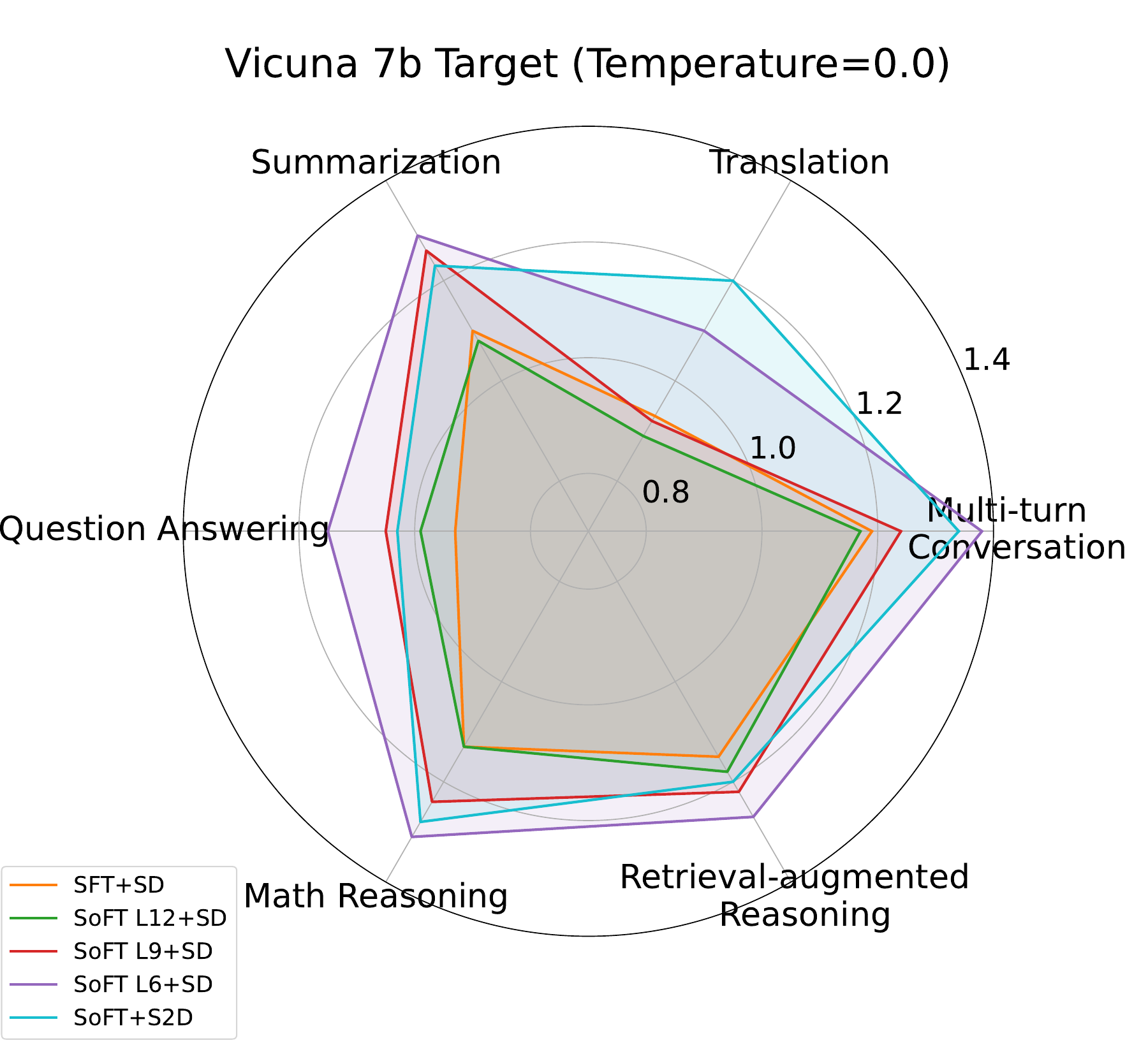}
    \end{subfigure}
    \hspace*{\fill}
    \begin{subfigure}{0.32\textwidth}
        \includegraphics[width=\linewidth]{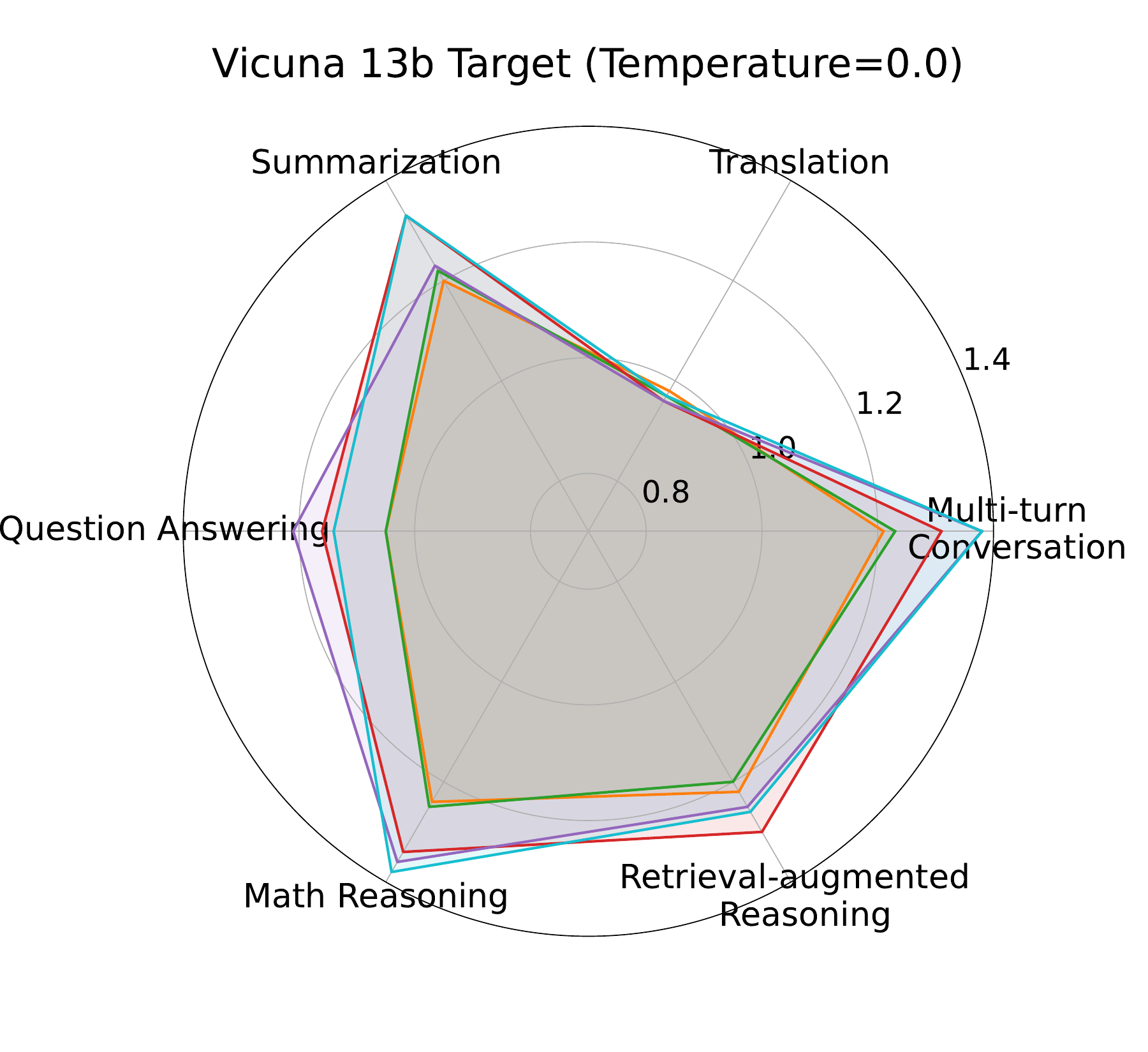}
    \end{subfigure}
    \hspace*{\fill}
    \begin{subfigure}{0.32\textwidth}
        \includegraphics[width=\linewidth]{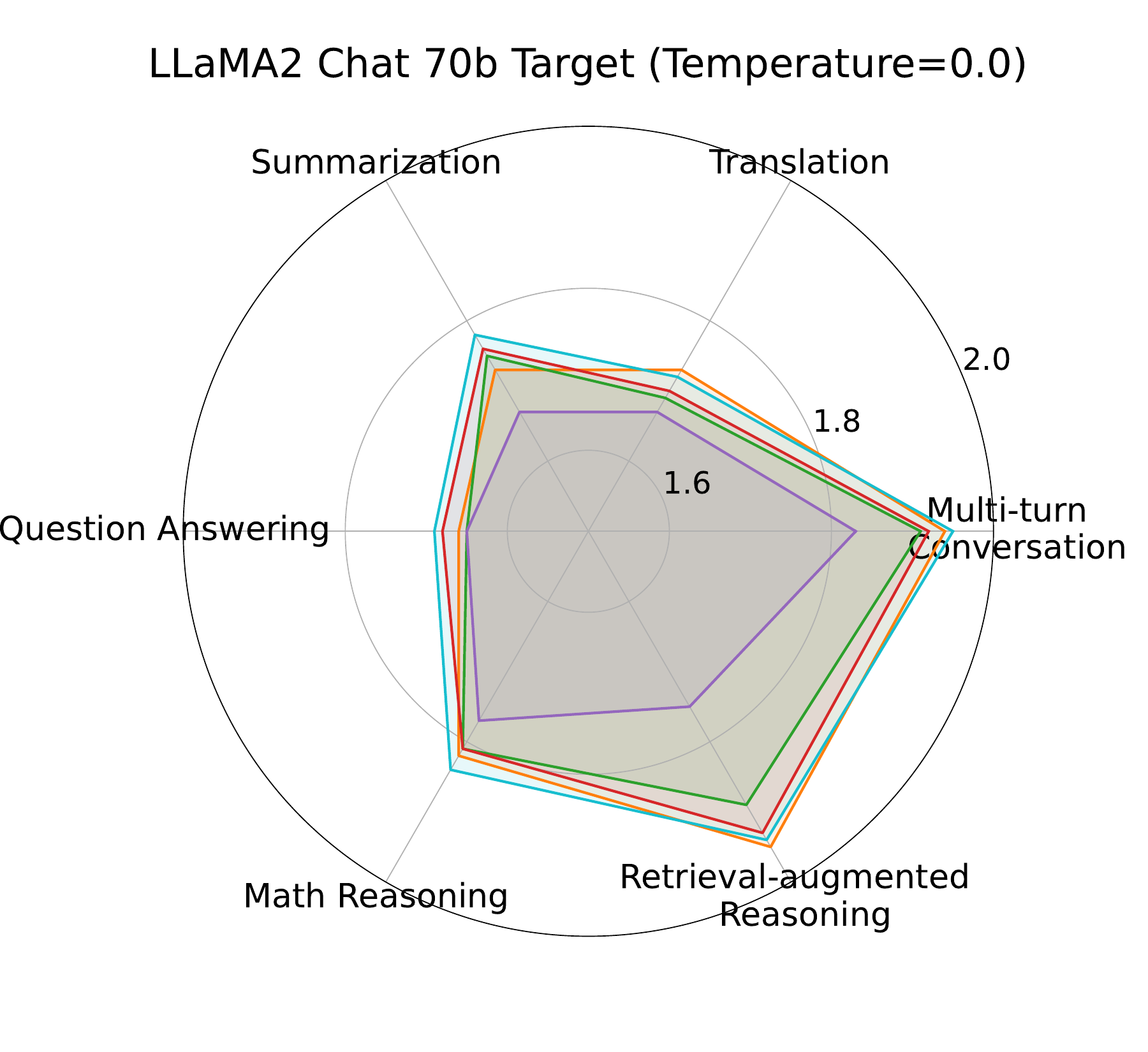}
    \end{subfigure}

    \caption{\small{Comparison among Speedup ratios of speculative and S2D methods on different domains on multiple targets.}} 
    \label{fig:spec-bench-main}
\end{figure*}

\paragraph{S2D vs baselines (speculative):}

Table \ref{tab:results} and Figure \ref{fig:spec-bench-main} depict the S2D performance compared to normal SD with different draft options on multiple targets and tasks. Taking Vicuna 7b and 13b as target models, S2D outperforms significantly SD with SFT and SoFT full draft models (12 layers). S2D has also higher speedup compared to SD with medium size (Layer 9) of SoFT model in many tasks like MT-Bench and GSM8K. SD with the smallest sub-model (Layer 6) of SoFT model outperforms S2D in most tasks due to the fact that the smallest sub-model of our draft architecture enjoys 1/2 latency of the full architecture, which plays an important role in the overall speedup in speculative decoding algorithm.

However, when it comes to larger target sizes like LLaMA Chat 70b, S2D roughly maintains the speedup of SD with SFT and SoFT full model architecture. While using SD with intermediate SoFT draft sub-models, especially smallest one (Layer 6), would cause significant drop in speedup compared to other draft options, indicating the importance of draft model capacity and Mean Accepted Tokens length factors when it comes to large targets inference acceleration.

Overall, depending on the necessity of lower draft latency or higher accepted token ratio, S2D performance demonstrates that our proposed approach can choose the sub-models accordingly to have the optimum speedup compared to other draft options in the architecture. The more details about the baselines performance on Spec-Bench can be found in Appendix \ref{appx:specbench}.

\subsection{Ablation Studies}

\subsubsection{Thresholds}
\label{sec:thresholds}
\begin{figure}[t]
    \centering
    \includegraphics[width=\linewidth]{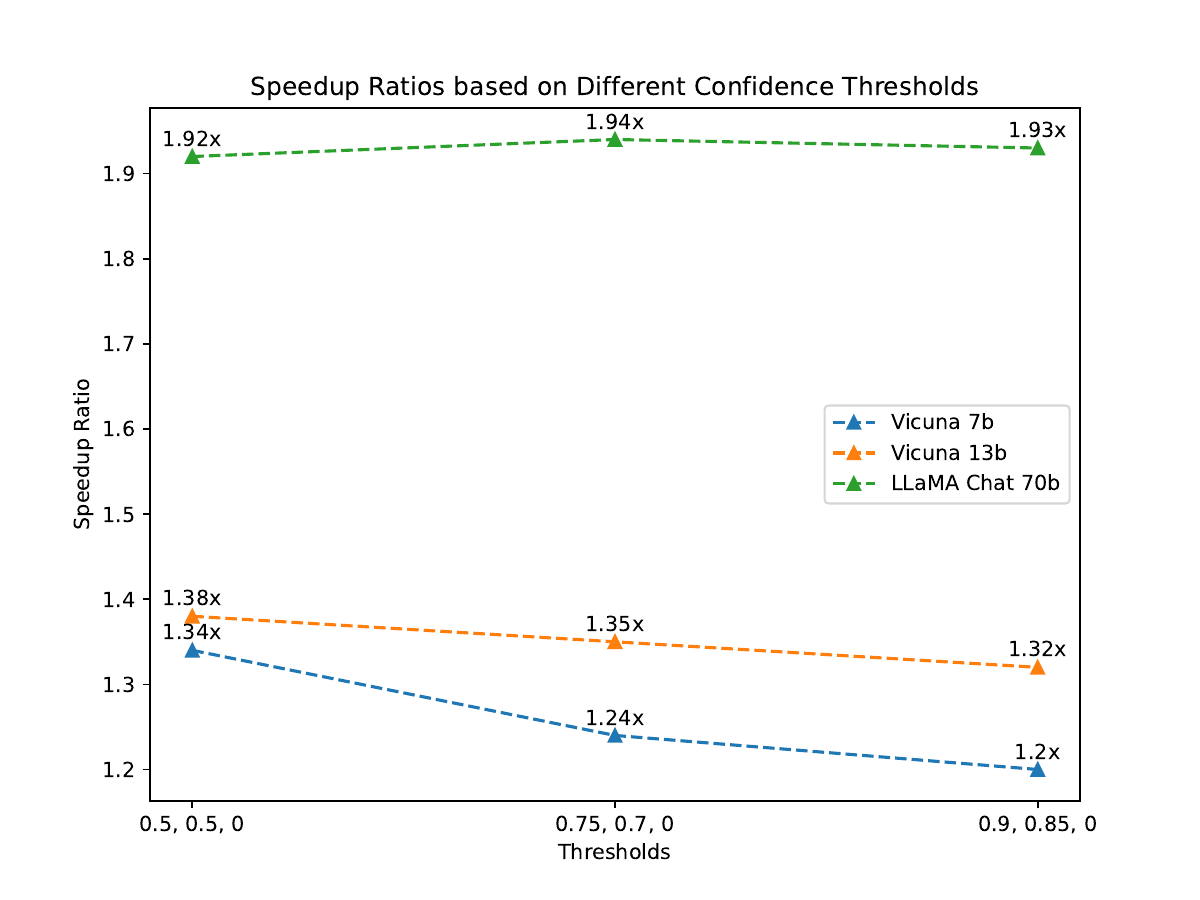}
    \caption{\small{Speedup ratios based on difference confidence thresholds in multiple targets on MT-Bench. In the thresholds axis labels, the first, second and third numbers represent the thresholds for the first draft sub-model (layer 6), the second draft sub-model (layer 9) and the last one (layer 12). The temperature was set to 0.0 in these experiments.}}
    \label{fig:thresholds}
\end{figure}
To find the optimum confidence thresholds for SoFT draft sub-models in S2D algorithm, we assess different thresholds sets to see the algorithm's performance in each scenario. Figure \ref{fig:thresholds} shows the performance comparison of different confidence thresholds. While in smaller target models, there is a tendency to choose the smaller draft sub-models by lowering their corresponding confidence thresholds to increase speedup, in the largest target model there is a need for higher draft model capacity therefore higher confidence thresholds result in optimum speedup. We fixed the best thresholds of each target model in all other experiments conducted with S2D algorithm in this paper.


\subsubsection{Impact of Attention Tree}

Unlike speculative decoding and sorted speculative decoding methods, recent approaches such as EAGLE \cite{li2024eagle}, Medusa \cite{cai2024medusa}, and Hydra \cite{ankner2024hydra} employ tree attention to simultaneously verify multiple candidate tokens. This addition complicates direct comparisons with other methods. Therefore, we also evaluated these methods without tree attention and provided the results in Table \ref{tab:results}. As expected, the speedup ratio significantly decreased for all. Interestingly, when tree attention is removed, Medusa's performance even falls below that of speculative decoding, specifically using layer 6 of the SoFT-trained sub-models.


\subsubsection{Impact of Feature Alignment}
Using the last layer hidden state representation of the target model to train a draft model has become a common approach in recent works \cite{cai2024medusa, ankner2024hydra}. More specifically, EAGLE \cite{li2024eagle} employs both features and tokens generated by the target model to train an one-layer draft. Supposed $t_i$ and $f_i$ are the token and hidden state feature generated by target LLM and $\hat{t}_i$ and $\hat{f}_i$ are the token and hidden state features generated by draft at position $i$, Eagle aligns hidden state features of draft and target by using L1 regression loss:  
\[
L_{\text{reg}} = \text{SmoothL1}(f_{i+1}, \text{Draft\_Model}(t_{2:i+1}, f_{1:i})).
\]
They also employ classification loss to directly optimize towards alignment of tokens:
\[
p_{i+2} = \text{Softmax}(\text{LM\_Head}(f_{i+1})),
\]
\[
\hat{p}_{i+2} = \text{Softmax}(\text{LM\_Head}(\hat{f}_{i+1})),
\]
\[
L_{\text{cls}} = \text{CrossEntropy}(p_{i+2}, \hat{p}_{i+2}).
\]
By integrating feature alignment (regression loss) and token alignments (classification loss), EAGLE's autoregression head is trained using the combined loss function: $L =  w_{\text{reg}} L_{\text{reg}} + w_{\text{cls}} L_{\text{cls}}.$

We conducted experiments studying the affects each feature and token alignment cause by setting different combination of $w_{\text{reg}}$ and $w_{\text{cls}}$ (Table \ref{tab:eagle_ablation}). We train eagle draft model based on a LLaMA2 13b target model fine-tuned on GSM8K train data. We found out that canceling the token alignment loss ($w_{\text{cls}}=0$) would not have a significant impact on the draft performance compared to the original setup used in Eagle paper ($w_{\text{reg}}=1$ and $w_{\text{cls}}=0.1$). On the other hand, setting $w_{\text{reg}}$ to 0 would cause a noticeable impact on the draft performance, dropping speedup from 2.73x to 1.61x. 
As we can see, feature alignment plays the main role in improving the Eagle draft performance while this is impractical in multi-target setting where the draft model needs to serve multiple target models anytime. 

\begin{table*}[t]
\setlength\extrarowheight{2pt}
\centering
\scalebox{0.75}{
\begin{tabular}{l|c|ccc}
\toprule
\multirow{2}{*}{\textbf{Model / Metric}} & \multirow{2}{*}{\textbf{Autoregressive}} & \multicolumn{3}{c}{\textbf{EAGLE}} \\
\cline{3-5} 
 &  & \textbf{w\_cls=0.1, w\_reg=1.0} & \textbf{w\_cls=0.0, w\_reg=1.0} & \textbf{w\_cls=1.0, w\_reg
 =0.0} \\
\midrule
\textbf{GSM8K EM} & 48.90 & 48.67 & 48.90 & 48.82 \\
\textbf{GSM8K Speedup Ratio} & 1.00x & 2.73x & 2.70x & 1.61x \\
\bottomrule
\end{tabular}}
\caption{
\small{Comparison of GSM8K exact match (EM) and speedup ratios for different decoding configurations. With Llama2 13B target model and 1-layer Eagle draft model.}
}
\label{tab:eagle_ablation}
\end{table*}

\subsubsection{Impact of Pre-training}  
Using pre-training draft models can be a possible direction to increase the acceptance ratio of draft tokens in speculative decoding algorithm. In this way, we repeated our experiments in a 
new setup where we replaced the first 12 layers of Vicuna 7b with Vicuna 160m, which is a LLaMA 160m checkpoint fine-tuned on ShareGPT dataset. LLaMA 160m is a small 12 decoder layer architecture pre-trained on C4 corpus. We also sorted fine-tuned the Vicuna 160m on ShareGPT with the same sub-models (Layer 6, 9 and 12). Table \ref{tab:pretraining_impact} shows the benefit of using S2D instead of regular SD for smaller target models (Vicuna 7b and 13b). Based on the results in Table \ref{tab:results}, fine-tuning an extracted 12-layer draft model can result in higher speedup compared to employing a similar pre-trained architecture, which can demonstrate the efficiency of our approach in terms of training resources.

\begin{table}[t]
 \setlength\extrarowheight{2pt}
 \centering
 \scalebox{0.6}{\begin{tabular}{l|cc|cc|cc}
  
  \toprule
    \multirow{3}{*}{\textbf{Method}} & \multicolumn{6}{c}{\textbf{Greedy (T = 0)}} 
   \\ \cmidrule{2-7}
    & \multicolumn{2}{c}{\textbf{Vicuna 7B}} & \multicolumn{2}{c}{\textbf{Vicuna 13B}} & \multicolumn{2}{c}{\textbf{LLaMA Chat 70B}} 
   \\ 
    & Speedup & MAT & Speedup & MAT & Speedup & MAT
   \\ \midrule
   Vicuna 160m + SD & 1.05$\times$ & 2.75 & 1.13$\times$ & 2.66 & 1.93$\times$ & 2.24 \\ 
   SoFT L6 + SD & 1.20$\times$ & 2.10 & 1.26$\times$ & 2.06 & 1.68$\times$ & 1.76 
   \\
    SoFT L9 + SD & 1.10$\times$ & 2.33 & 1.21$\times$ & 2.29 & 1.79$\times$ & 1.96 
   \\
    SoFT L12 + SD & 1.00$\times$ & 2.67 & 1.07$\times$ & 2.58 & 1.90$\times$ & 2.19 
   \\
     SoFT + S2D & 1.17$\times$ & 2.54 & 1.28$\times$ & 2.44 & 1.91$\times$ & 2.16 
   \\
   \bottomrule
 \end{tabular}}
 \caption{
  \small{Overall Speedup and Mean Accepted Tokens lenght (MAT) on MT-Bench dataset. The SoFT training in this experiment was initialized with Vicuna 160m checkpoint.}
 }
 \label{tab:pretraining_impact}
\end{table}

\subsubsection{Target Model Training}

Foundation models like GPT-4, Gemini \citep{team2023gemini}, and the Claude family \citep{Claude3} are trained on vast datasets, enabling them to perform well across various tasks. Nonetheless, for specialized tasks where the model has limited exposure during pre-training, domain adaptation through fine-tuning leads to superior performance in downstream tasks \citep{liu2024panda}. This section evaluates the impact of fine-tuning both the target and draft models for specific tasks on the speed gains achievable using our proposed sorted speculative decoding method. In this way, we fine-tune a Llama 13B model and also sorted fine-tuned the extracted 12-layer draft on GSM8k, a mathematical reasoning dataset. Results shown in Table \ref{tab:target_model_training} demonstrate that fine-tuning both target and draft on the same dataset, due to their improved alignment, results in a 1.14$\times$ speed increase at inference time.

\begin{table}
\setlength\extrarowheight{2pt}
\centering
\scalebox{0.75}{
\begin{tabular}{l|ccc}
\toprule
& \multicolumn{3}{c}{\textbf{GSM8K}} \\
\midrule
\multirow{2}{*}{\textbf{Model}} & \multicolumn{3}{c}{\textbf{Auto-regressive Decoding}} \\ 
& \textbf{Trained Target}  & \textbf{Speedup} & \textbf{Accuracy}  \\
\midrule
Llama2 13B & \cmark  & 1$\times$ & 48.97 \\
Llama2 13B & \xmark & 1$\times$ & 28.7 \\
\midrule
\multirow{2}{*}{\textbf{Draft Model}} & \multicolumn{3}{c}{\textbf{Sorted Speculative Decoding}} \\
& \textbf{Trained Target}  & \textbf{Speedup} & \textbf{Accuracy}  \\
\midrule
S2D - SoFT  & \cmark  & \textbf{1.53$\times$} & 48.97 \\
S2D - SoFT  & \xmark  &  1.38$\times$ & 28.7 \\
\bottomrule
\end{tabular}
}
\caption{\small{Comparison of speedup between two different settings: 1) training the target model on the downstream task 2) using the vanilla pre-trained model}}
\label{tab:target_model_training}
\end{table}

\section{Conclusion}

In this paper, we present a method based on the SoFT training of a draft model to overcome a significant limitation of traditional speculative decoding methods, where each target model necessitates a uniquely trained draft model. Through comprehensive experimentation, we demonstrate that by using a same SoFT-trained draft model with varying thresholds for sub-models, we achieve an average speedup ratio of \textbf{1.55} for target models with parameters ranging from 7B to 70B. Moreover, our method surpasses vanilla speculative decoding across all target models, highlighting its effectiveness.

\section{Limitation}

Although our proposed method accommodates a diverse range of target models through its SoFT-trained sub-models for token prediction, it introduces sub-model thresholds as a new hyperparameter that requires tuning. Nonetheless, this requirement is considerably much simpler to undertake compared to the alternative approaches that involve training separate draft models for each target. However, based on our experiments larger thresholds work better with larger target models and smaller thresholds should be used with smaller target models

Additionally, this paper primarily focuses on a specific setting where a draft model is trained using SoFT with three sub-models. However, exploring a different number of sub-models from different layers could offer deeper insights into our methodology. We identify these comparative studies as potential future work.

\bibliography{custom}

\newpage 

\section{Appendix}

\subsection{Hyperparameters}
\label{appx:hyperparameters}
The batch size during the ShareGPT training experiments (SFT and SoFT) was 8 and gradient accumulation steps was set to 16. We used 4 NVIDIA V100 GPUs for training experiments. Each inference experiment on Spec-bench was done on 2 NVIDIA V100 GPUs, except the experiments for LLaMA Chat 70b where we used 8 GPUs to avoid memory issues.

\subsection{Spec-bench Results}
\label{appx:specbench}

\begin{figure*}[t!] 
    \begin{subfigure}{0.5\textwidth}
        \includegraphics[width=\linewidth]{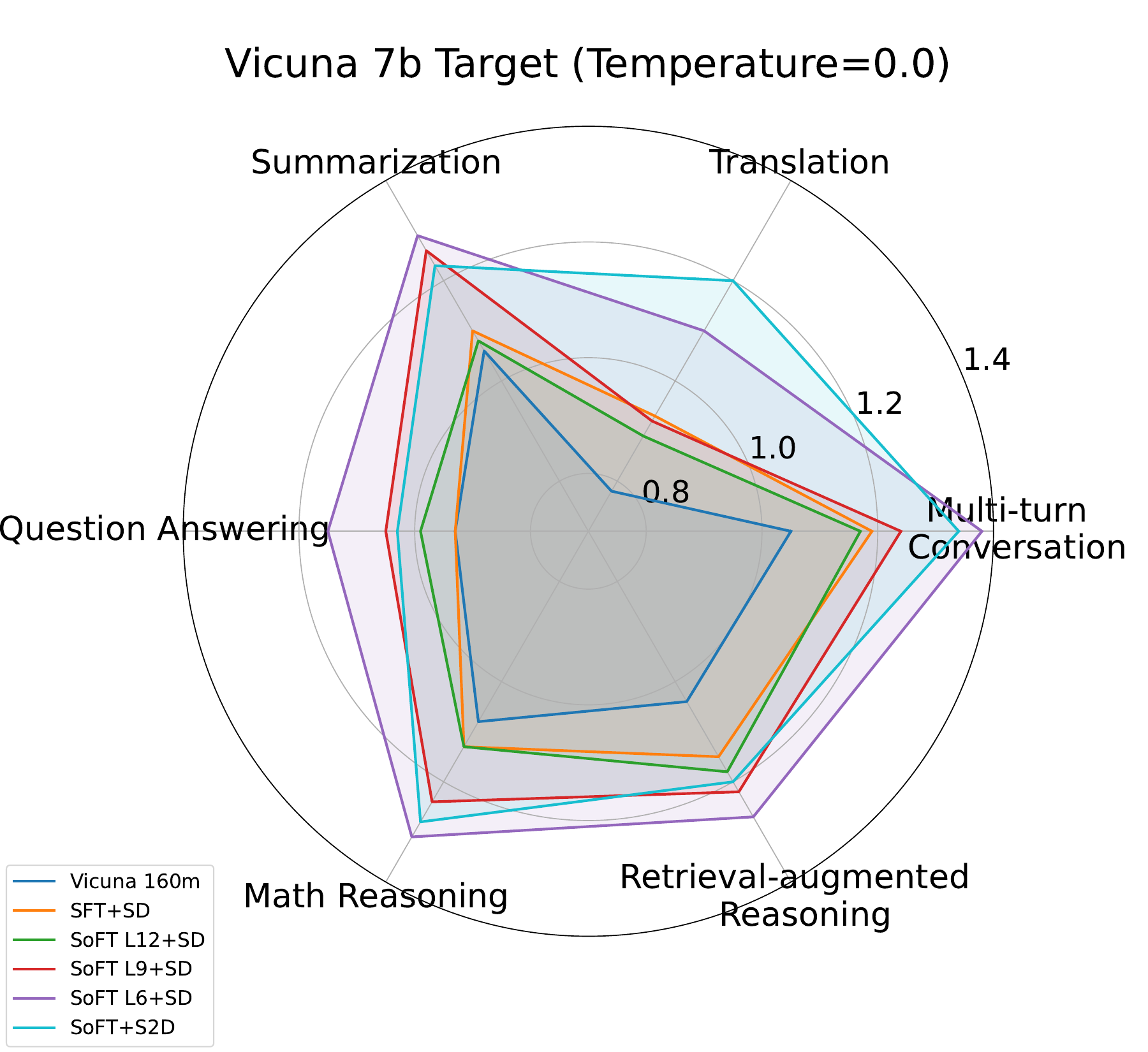}
    \end{subfigure}
    \hspace*{\fill}
    \begin{subfigure}{0.5\textwidth}
        \includegraphics[width=\linewidth]{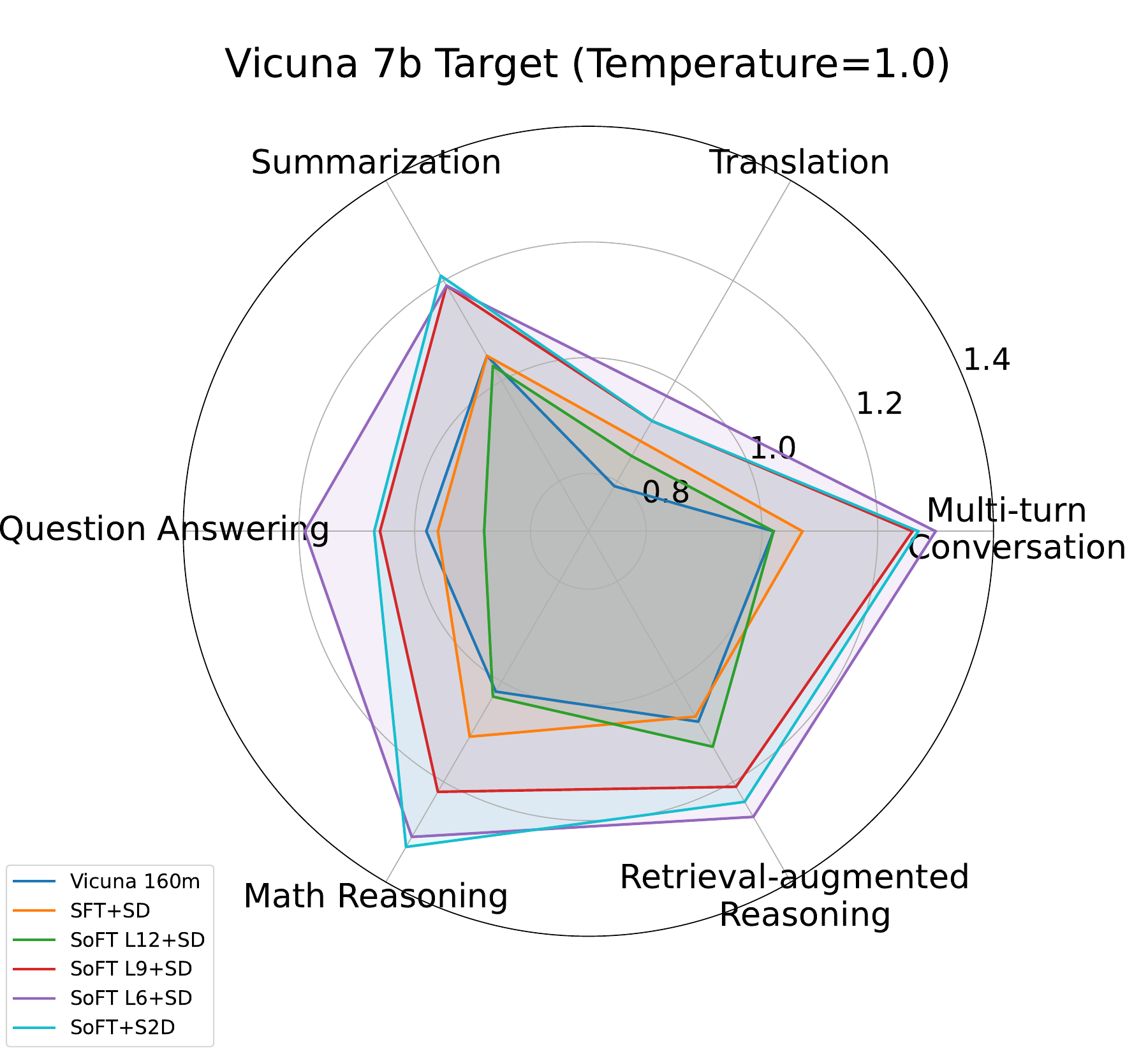}
    \end{subfigure}
    \hspace*{\fill}
    \begin{subfigure}{0.49\textwidth}
        \includegraphics[width=\linewidth]{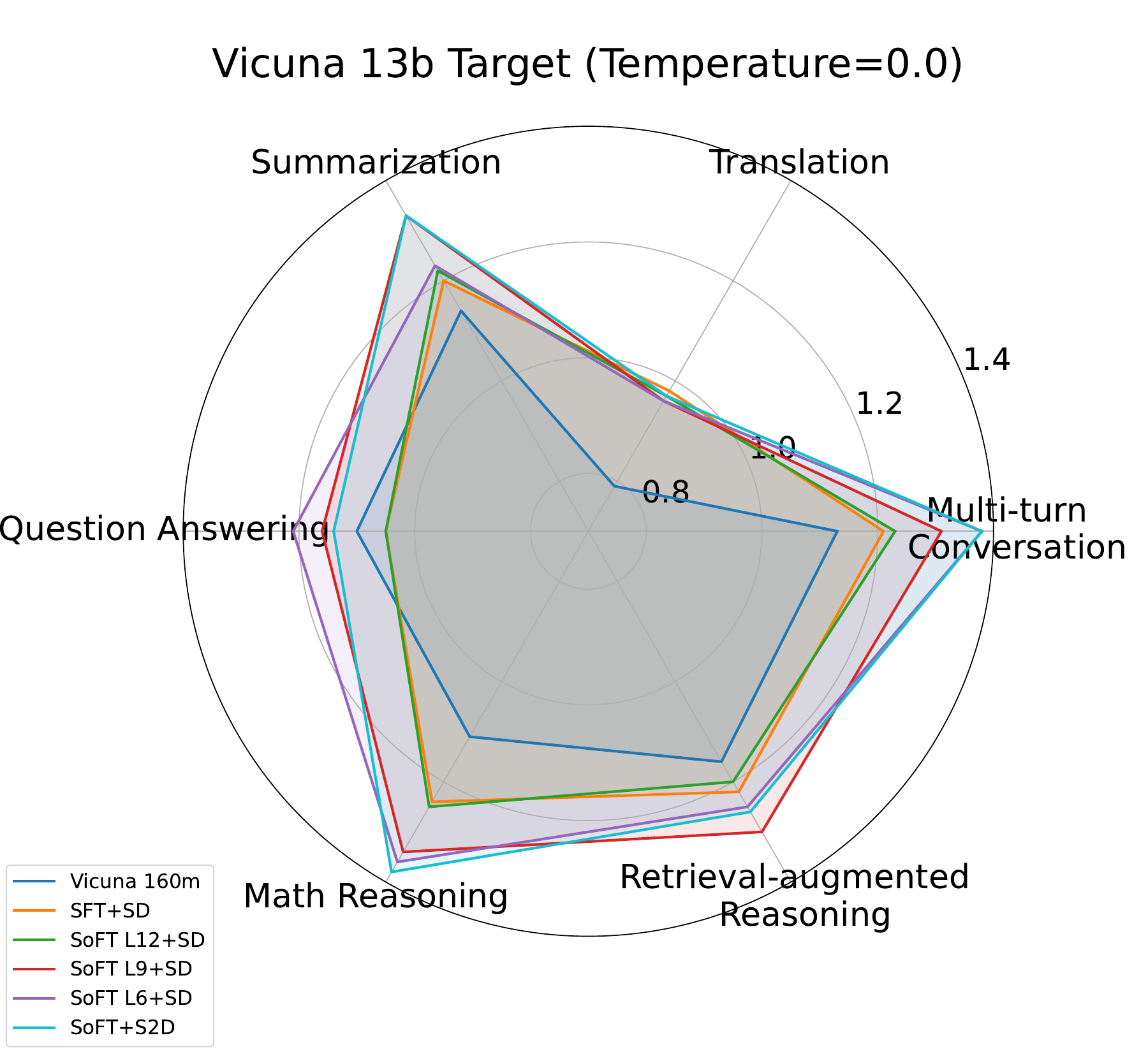}
    \end{subfigure}
    \begin{subfigure}{0.49\textwidth}
        \includegraphics[width=\linewidth]{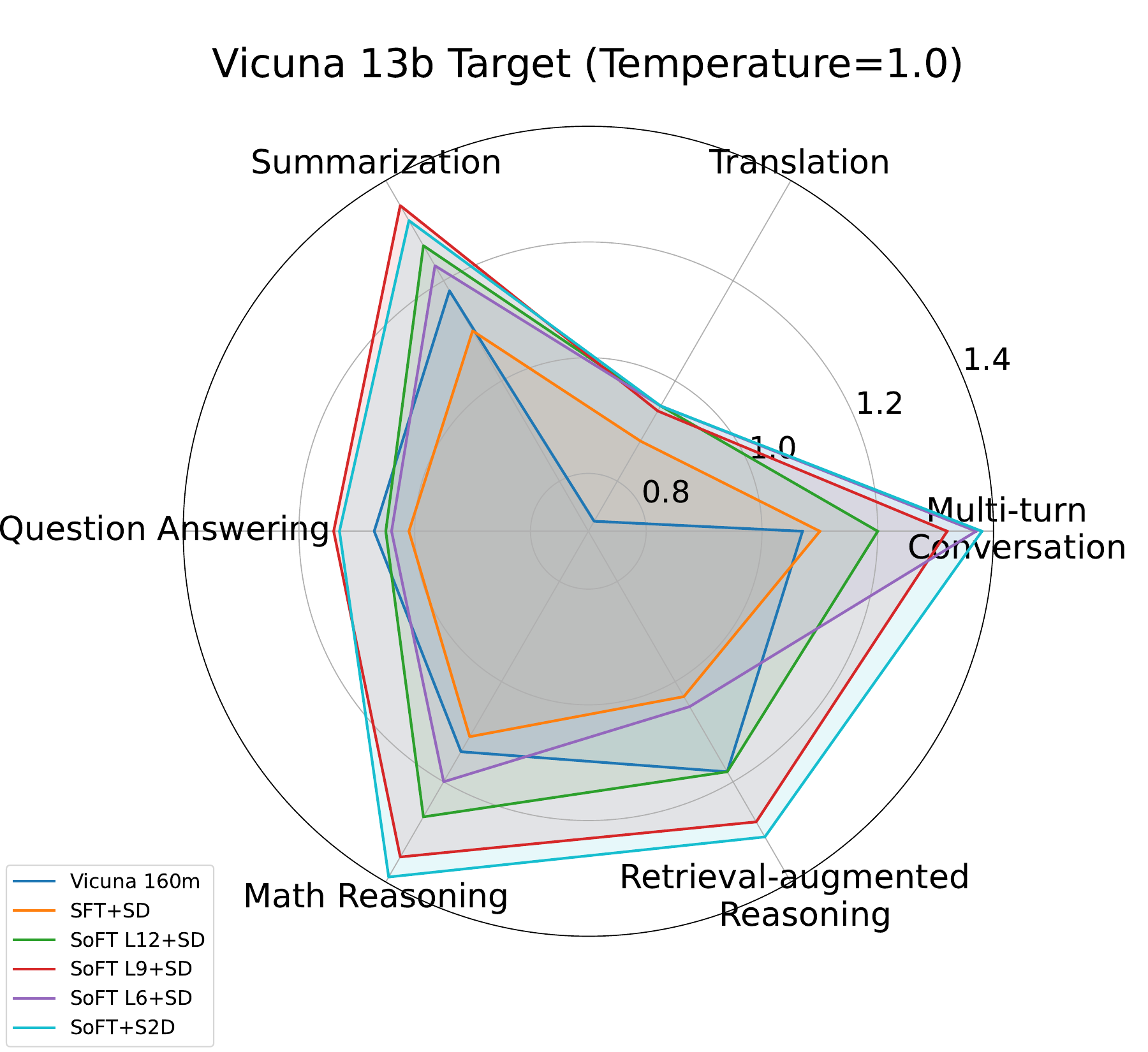}
    \end{subfigure}
    \begin{subfigure}{0.49\textwidth}
        \includegraphics[width=\linewidth]{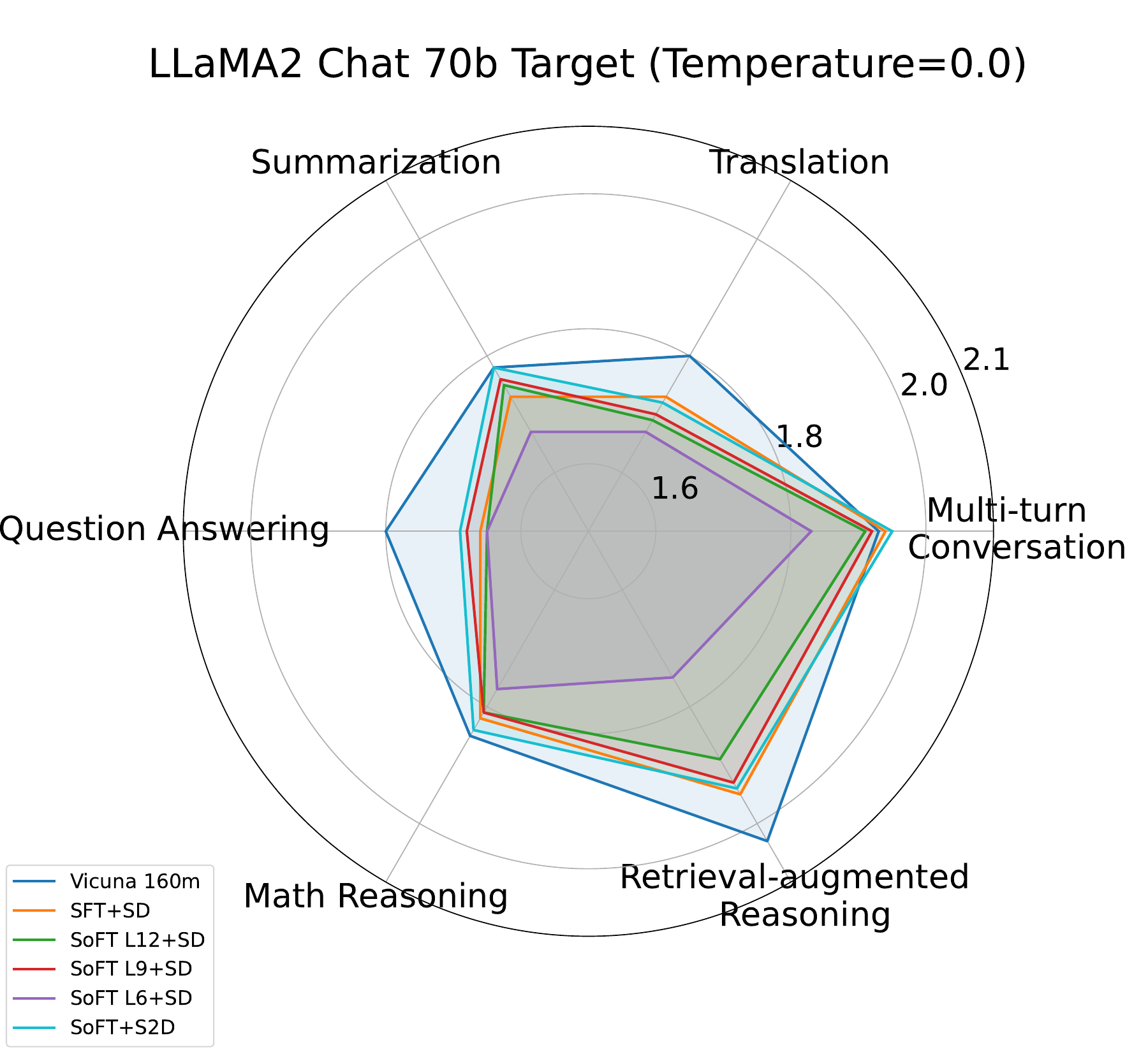}
    \end{subfigure}
    \hspace*{\fill}
    \begin{subfigure}{0.49\textwidth}
        \includegraphics[width=\linewidth]{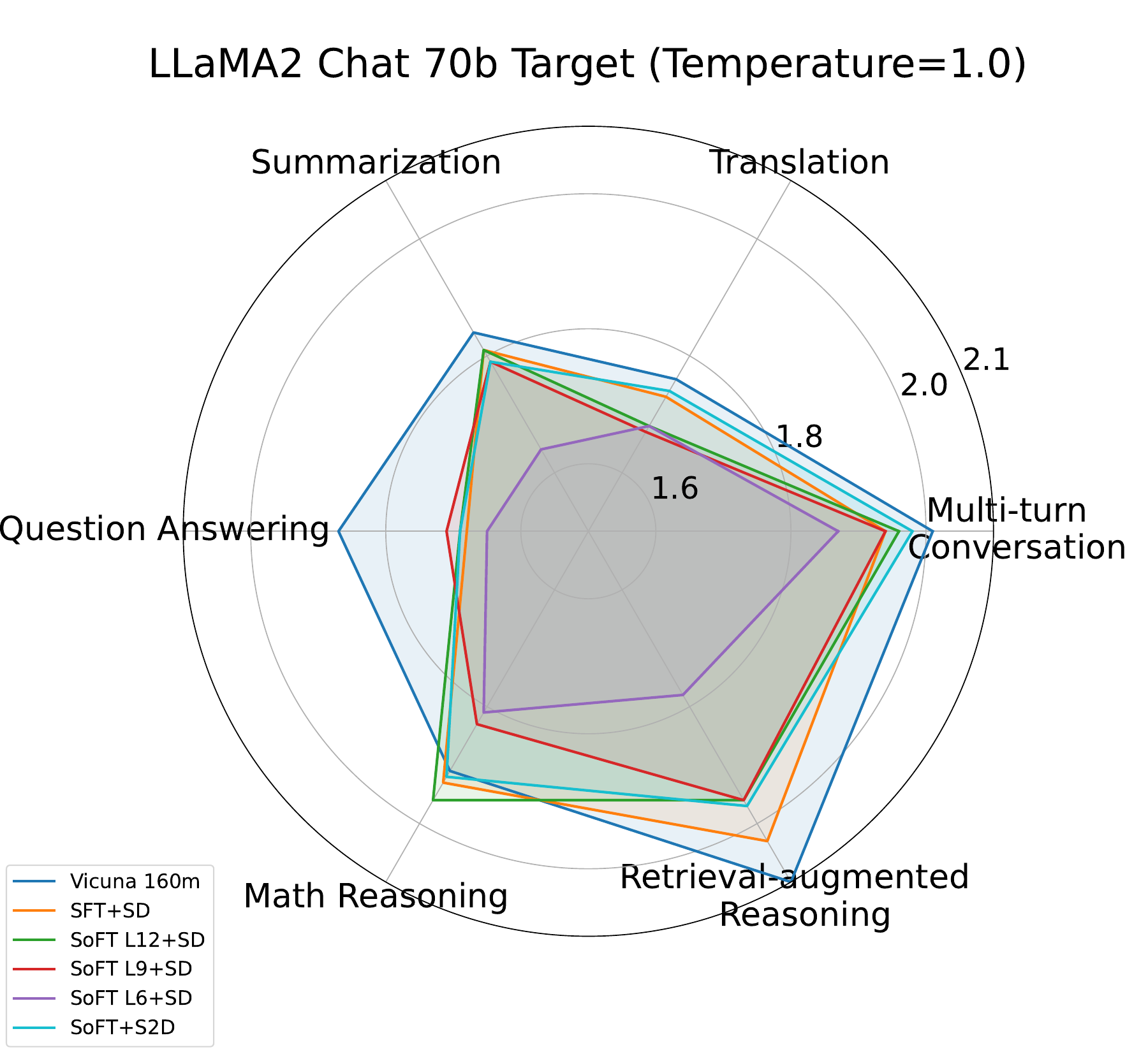}
    \end{subfigure}

    \caption{Comparison among Speedup ratios of speculative and S2D methods on different domains on multiple targets.} 
    \label{fig:spec-bench-appendix}
\end{figure*}

Figure \ref{fig:spec-bench-appendix} shows the speedup ratios of S2D versus speculative decoding with different draft options, including Vicuna 160m pre-trained. As S2D demonstrate superior performance than speculative decoding with the same size draft models (12 layers) in smaller target sizes (Vicuna 7B and 13B), once we get to larger target sizes, larger drafts outperform smaller ones (Layer 6 and 9) in speculative decoding algorithm. This can demonstrate the importance of capacity and accepted tokens length in larger target sizes. However, S2D can maintain a similar performance to speculative decoding with the same draft size even in case of using a target with 70B size.

\end{document}